\documentclass[runningheads]{llncs}
\usepackage{graphicx}
\usepackage{algorithm}
\usepackage{xcolor}
\usepackage[]{algpseudocode}
\usepackage{subfig}
\usepackage[hyphens]{url}
\begin{document}
\title{Integrating Unsupervised Clustering and Label-specific Oversampling to Tackle Imbalanced Multi-label Data\thanks{This work has emanated from research supported in part by TCG Crest, Kolkata, India and a grant from Science Foundation Ireland under Grant number [16/RC/3835]. For the purpose of Open Access, the author has applied a CC BY public copyright licence to any Author Accepted Manuscript version arising from this submission.}}
\titlerunning{UCLSO}
%
\author{Payel Sadhukhan\inst{1}\orcidID{0000-0001-7795-3385} \and Arjun Pakrashi \inst{2}\orcidID{0000-0002-9605-6839} \and Sarbani Palit \inst{3}\orcidID{0000-0002-4105-6452} \and Brian Mac\ Namee \inst{2}\orcidID{0000-0003-2518-0274}}
\authorrunning{P. Sadhukhan et al.} 
%
\institute{TCG Crest Kolkata, India \\ \email{payel0410@gmail.com}\and School of Computer Science, University College Dublin, Ireland \email{\{arjun.pakrashi,brian.macnamee\}@ucd.ie} \and Indian Statistical Institute Kolkata, India\\ \email{palitsarbani@gmail.com}}
\maketitle              
\begin{abstract}
There is often a mixture of very frequent labels and very infrequent labels in multi-label datatsets. This variation in label frequency, a type class imbalance, creates a significant challenge for building efficient multi-label classification algorithms. In this paper, we tackle this problem by proposing a minority class oversampling scheme, UCLSO, which integrates Unsupervised Clustering and Label-Specific data Oversampling. Clustering is performed to find out the key distinct and locally connected regions of a multi-label dataset (irrespective of the label information). Next, for each label, we explore the distributions of minority points in the cluster sets. Only the minority points within a cluster are used to generate the synthetic minority points that are used for oversampling. Even though the cluster set is the same across all labels, the distributions of the synthetic minority points will vary across the labels. The training dataset is augmented with the set of label-specific synthetic minority points, and classifiers are trained to predict the relevance of each label independently. Experiments using 12 multi-label datasets and several multi-label algorithms show that the proposed method performed very well compared to the other competing algorithms.
\end{abstract}

\section{Introduction}\label{sec:introductiton}
In a multi-label dataset a single datapoint is associated with more than one relevant label. This type of data is obtained naturally from real-world domains like text ~\cite{text,Godbole}, bioinformatics ~\cite{gene1}, video \cite{video}, images ~\cite{Boutell,image3,Guo-Image} and music ~\cite{music1}. We denote a multi-label dataset as $\mathcal{D}=\{(\mathbf{x}_{i},\mathbf{y}_{i}),\mathit{i=1,2,\dots,n}\}$. Here, $\mathbf{x}_i$ is the $i^{th}$ input datapoint in $d$ dimensions, and $\mathbf{y}_{i}=\{y_{i1},y_{i2},\dots,y_{iq}\}$ is the corresponding label assignment for $\mathbf{x}_i$ among the possible $q$ labels. $y_{ik}$ indicates if the $k^{th}$ label is applicable (or \emph{relevant}) for the $i^{th}$ datapoint:  $y_{ik} = 1$ denotes that the $k^{th}$ label is relevant to $x_i$, and $y_{ik} = 0$ denotes that the $k^{th}$ label is not applicable, or is \emph{irrelevant}, to $x_i$. The target of multi-label learning is to build a model that can correctly predict all of the relevant labels for a datapoint $\mathbf{x}_{i}$.


Multi-label datasets are often found to possess an imbalance in the representation of the different labels---some labels are relevant to a very large number of datapoints while other labels are only relevant to a few. We can consider this an example of the class imbalance problem common in binary classification problems if we consider the relevance of each label to be analogous to a binary classification. Often, the labels in a multi-label dataset have widely varying degrees of imbalance and this is a challenging aspect of building multi-label classification models. 


Addressing label imbalance to improve multi-label classification is an active field of research, and several methods have been proposed to address this problem \cite{cocoa,daniels,liu2019}. There is, however, room for significant improvement. Label-specific oversampling can be a solution to address the issue of varying label imbalances in multi-label datasets. In this light, we propose UCLSO, which integrates \emph{\textbf{U}nsupervised \textbf{C}lustering and \textbf{L}abel \textbf{S}pecific data \textbf{O}versampling}. The essence of the UCLSO approach is to integrate information about the proximity of points and their label-specific class-memberships to solve the issue of class imbalance in multi-label datasets. In this work, i) synthetic minority points are generated from local data clusters which are obtained from unsupervised clustering of the feature space, and ii) the cardinality of the label-specific oversampled minority set obtained from a local cluster will depend on the cluster's  share of minority datapoints for that label. In effect, the method oversamples the minority class by focusing on per-cluster local distributions of the minority datapoints to maintain the local minority distribution ratio.
The key highlights of our work are,
\begin{itemize}
\item We propose UCLSO, a new minority class oversampling method for multi-label datasets, which generates synthetic minority datapoints specifically in the minority regions of the input space.
\item UCLSO preserves the intrinsic class distributions of the local clusters in order to avoid generating synthetic minority datapoints in the majority region, or as outliers in the input space.
\item UCLSO ensures that the number of synthetic minority points added to a region is proportionate to the original minority density in that region.
\item In UCLSO, datapoints belonging to individual clusters (consistent across the labels) have distinct label relevance which vary across the different labels. We integrate this label-specific information along with the information from the previous step to obtain sets of \textit{label-specific synthetic minority points}.
\item An empirical study involving 12 well-known real-world multi-label datasets and nine competing methods indicates that UCLSO shows promising results and is able to perform better, in general, than the competing methods. 
\end{itemize}
The remainder of the paper is structured as follows. Section \ref{sec:related_work} discusses the relevant existing work in the multi-label domain. In Section \ref{sec:proposedmethod} we first describe the motivations of our approach and then present the steps of the proposed UCLSO algorithm. The experiment design is described in Section \ref{sec:experiments} and the results of the experiments are discussed in Section \ref{sec:results}. Finally, Section \ref{sec:conclusion} concludes the paper and discusses some directions for future work.

\section{Related Work}\label{sec:related_work}
Existing multi-label classification methods are principally classified into two types: i) \emph{Problem transformation} methods that modify the multi-label dataset in different ways such that it can be used with existing multi-class classification algorithms \cite{rakel,Calibrated,cc,review-zhang}, and ii) \emph{Algorithm adaptation} approaches that modify existing machine learning algorithms to directly handle multi-label datasetets \cite{mlknn,bpmll-large,bpmll,review-zhang}.


Multi-label algorithms can also be categorised based on if and how they take label associations into account, which allows algorithms to be categorised as: i) \textit{first-order}, ii) \textit{second-order} or iii) \textit{higher-order} approaches based on the number of labels that are considered together to train the models. First order approaches do not consider any label association and learn a classifier for each label independently of all other labels \cite{mlknn,tanaka,zhangbr}. In second order methods, pair-wise label associations are explored to achieve enhanced learning of multi-label data \cite{Pairwise,Calibrated}. Higher order approaches considering associations between more than two labels \cite{Boutell}. A number of diversified techniques have facilitated higher order label associations through interesting schemes including classifier chains \cite{bayes-cc,PMCC}, RAkEL \cite{rakel}, random graph ensembles \cite{random-graph}, DMLkNN \cite{younes2008multi}, IBLR-ML+ \cite{cheng2009combining}, and Stacked-MLkNN \cite{Pakrashi}.

In recent years, data transformation has been a popular choice for handling multi-label datasets. The two principal ways of data transformation in multi-label domain are: i) feature extraction or selection, and ii) data oversampling or undersampling. One of the earliest applications of  feature extraction in multi-label learning was through LIFT \cite{lift-pami}, which brought significant performance improvements. Most feature selection or extraction methods select a label-specific feature set for each label to improve the discerning capability of the label specific classifiers. Subsequently, a number of different feature selection and extraction approaches have been proposed \cite{jfsc,xu-feature,xu2016multi,feature2}. Recently, the class imbalance problem in multi-label learning has received more interest from the researchers. One common approach to handling imbalance is to balance the cardinalities of the relevant and irrelevant classes for each label. One way of achieving this is through the removal of points from the majority class of each label-- for example using random undersampling \cite{irus} or tomek-link based undersampling \cite{tomek-undersampling}. Another way to achieve this is by adding synthetic minority points to the minority class  \cite{liu2019,SADHUKHAN,CHARTE}. Although this approaches have been shown to be effective there is still a lot of room for improvement.



\section{Unsupervised Clustering and Label Specific data Oversampling (UCLSO)}\label{sec:proposedmethod}
In this section we discuss the motivation and then present the proposed approach: Unsupervised Clustering and Label-Specific data Oversampling (UCLSO).

\subsection{Motivation}\label{sec:motivation}

\begin{figure}[!b]
\centering
\includegraphics[width=0.96\textwidth]{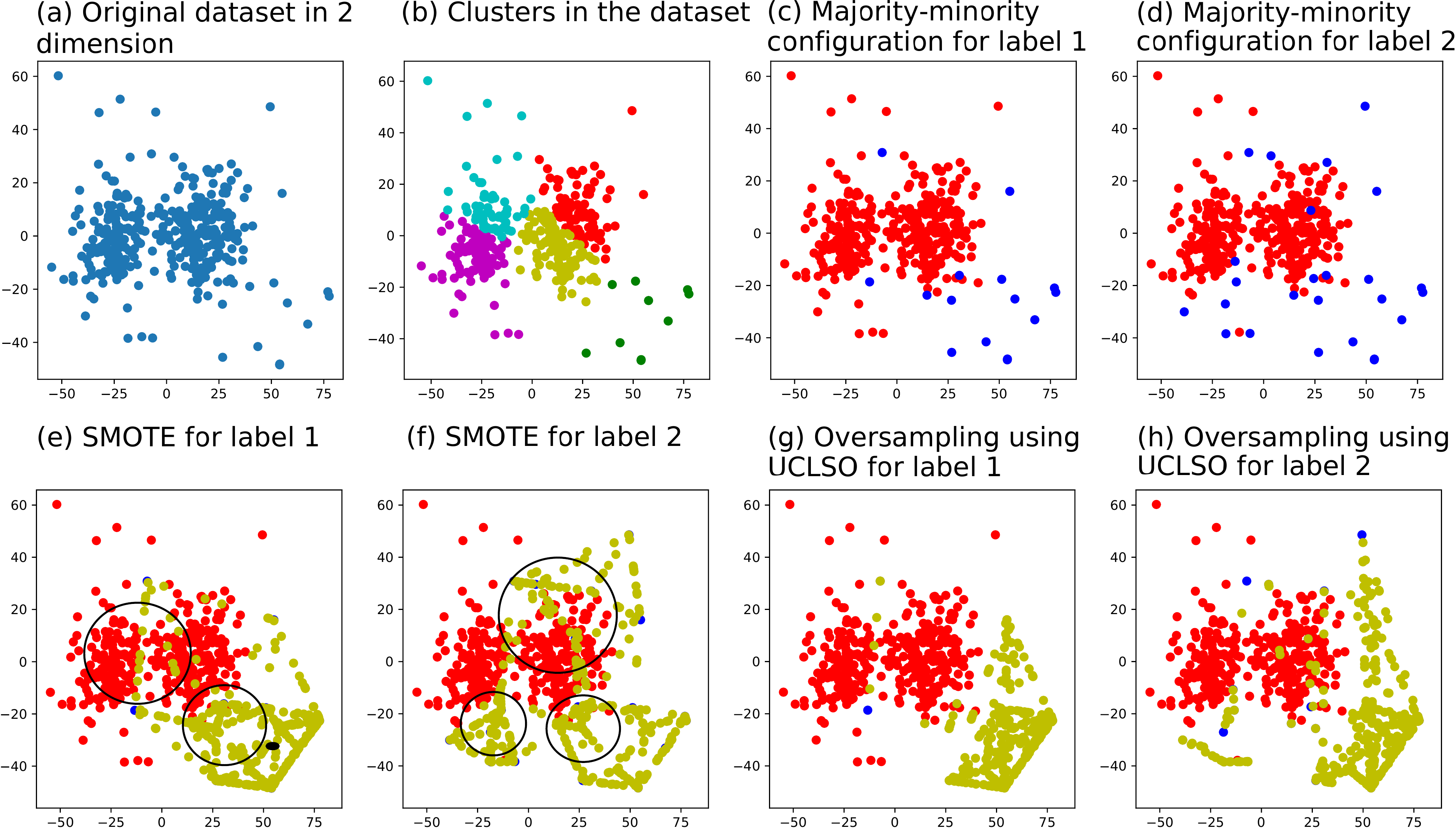}
\caption{A toy dataset illustrating the problems with oversampling, and how UCLSO addresses them}\label{fig:motivation_fig}
\end{figure}

Let us consider the two-dimensional toy dataset with two labels (1 and 2) shown in Figure \ref{fig:motivation_fig}(a). The imbalance ratios of labels 1 and 2 in this dataset are 24.7 and 14.4 respectively. Figure \ref{fig:motivation_fig}(b) shows 5 clusters in this datatset found using k-means. In Figures \ref{fig:motivation_fig} (c) and (d), we mark the points with respect to their label-specific class memberships. The colours red and blue  indicate majority and minority class points respectively. Data pre-processing via minority class oversampling is a popular choice to tackle the issue of imbalance in imbalanced datasets \cite{He}. In a multi-label dataset, due to spatial and quantitative variation of class-memberships across the labels, we need to label specific oversampling. Figures \ref{fig:motivation_fig} (e) and (f) show the label-specific SMOTE-based \cite{smote} oversampling (synthetic points in yellow) for label 1 and label 2 respectively. It can be seen that SMOTE oversamples the synthetic minority points in majority regions on a number of occasions for both labels 1 and 2 (highlighted by black circles in Figures \ref{fig:motivation_fig} (e) and (f)).

In order to achieve effective learning of a dataset, we need to prevent the majority space encroachment during oversampling. We tackle this issue by clustering (using k-means) the feature space. Clustering the dataset will give us k localized subspaces. Oversampling only within each cluster can prevent the majority class encroachment. 

This work is motivated by an effort to balance the cardinalities of the minority and majority classes of the labels without encroaching on the majority class spaces, as well as an effort to preserve the underlying distribution of the datapoints.

As indicated in Figures \ref{fig:motivation_fig} (e) and (f), a generic oversampling for all labels will not be fruitful as different labels have different quantitative and spatial distribution of the minority points. The are two aspects we need to keep in mind. i) Where should we perform the oversampling? To answer this, we cluster the feature space in an unsupervised manner (only the feature attributes of the points are taken into account). ii) If there is more than one subspace in which to perform oversampling, how much should we oversample in each subspace? We look into the distribution of the minority points (label-specific) in the clusters to decide this. The degree of label-specific oversampling in a cluster should be proportional to its original minority class distribution for that label. Figures \ref{fig:motivation_fig} (g) and (h) show the oversampling on labels 1 and labels 2 through the proposed method UCLSO. The degree of encroachment in the majority class region is much less for UCLSO compared to SMOTE. The next section will present the proposed UCLSO method in detail.

\subsection{Approach}

\begin{algorithm}[ht]
\caption{UCLSO}\label{alg:uclso}
\begin{algorithmic}[1]
\Procedure{UCLSO}{$\mathcal{D},k$}\Comment{$\mathcal{D}$: Training dataset, $k$: k-means clusters}
\State $\{C_1, C_2, \ldots, C_k\}$ = k\_means ($\{\mathbf{x}_i|1\le i \le n\}, k$) \Comment{Cluster input space}
\For{$l \in 1,2,\ldots,q$}
\State $\mathcal{S}_l = \{\}$
\For{$p \in  1,2,\ldots k$}
\State $n_{lp} = $ number of original minority points for label $l$ in $C_p$
\State \rule{0.1\textwidth}{1px}\Comment{Find the synthetic minority instance shares of each cluster}
\State $min_{l} = \{\mathbf{x}_i|\forall_i y_{il} = 1\}$
\State $maj_{l} = \{\mathbf{x}_i|\forall_i y_{il} = 0\}$
\State $syn_{lp} = \lceil n_{lp}\times\frac{|maj_{l}|-|min_{l}|}{|min_{l}|} \rceil,\hspace{0.2cm} p = 1, 2, \ldots, k$

\State \rule{0.1\textwidth}{1px}\Comment{Generate synthetic points}
\For{$j\in 1,2,\ldots,syn_{lp}$}
\State $\mathbf{u}_p$ is selected randomly from $C_p$
\State $\mathbf{v}_{p}$ is $\mathbf{u}_p$'s randomly selected nearest neighbor in $C_p$
\State $r \in (0, 1)$ selected randomly
\State $\mathbf{s}_{pj}^{(l)} = \mathbf{u}_p + (\mathbf{v}_p - \mathbf{u}_p) \times r$ \Comment{$j^{th}$ synthetic pt. for label $l$ from $C_p$}
\State $\mathcal{S}_{l} = \mathcal{S}_{l} \bigcup \{(\mathbf{s}_{pj}^{(l)},y_{pj}^{(l)}=1)\}$
\EndFor
\EndFor
\State $\mathcal{A}_l = \mathcal{D} \cup \mathcal{S}_l$
\Comment{Augment original data}
\EndFor\label{euclidendwhile}
\State \textbf{return} $\{\mathcal{A}_{l}|l=1,2,\ldots,q\}$\Comment{Per label augmented synthetic datasets}
\EndProcedure  
\end{algorithmic}
\end{algorithm}

Following the motivation in the previous section, in this work we propose a first-order  oversampling method for multi-label classification datasets, UCLSO, that handles class-imbalance for each label independently. 

The main idea of this oversampling method of minority points is to keep the synthetic minority points concentrated within the minority region of the input space. This is done to introduce more synthetic minority points in the minority regions of the input space in a guided fashion, while avoiding introduction of synthetic minority points in non-minority regions. This ideally should improve the minority label area representation, which will help in classifier algorithm training to define a better decision boundary for the specific imbalanced label. 


A common approach to generating synthetic minority datapoints is to select two points within a neighbourhood and then generate a synthetic point by interpolation at a random location between them. For a label with a high imbalance ratio and sparsely distributed minority points, the neighbours from the same class for this label can lie far apart. Consequently, the neighbourhood can encompass a large volume of feature space. Therefore, oversampling in such a manner may lead to the generation of synthetic minority points which end up in the majority region of the input space.

To tackle this issue, we partition the original points into $k$ clusters $\{C_1, C_2, \ldots, C_k\}$, based only on the input space inter-point Euclidean distances. We use the k-means algorithm to perform this clustering. After clustering the datapoints, for each cluster $C_p$, we randomly select $\mathbf{u}_p$, a minority point from the cluster, and $\mathbf{v}_p$, which is a randomly chosen nearest neighbour of $\mathbf{u}_p$ in $C_{p}$. We compute the synthetic minority point by interpolation at a random location of the direction vector connecting $\mathbf{u}_p$ and $\mathbf{v}_p$. The synthetic point is computed as follows

\begin{equation}
\mathbf{s}_{pj}^{(l)} = \mathbf{u}_p + (\mathbf{v}_p - \mathbf{u}_p) \times r
\end{equation}

\noindent where $s_{pj}^{(l)}$ is the $j$th synthetic datapoint generated in cluster $C_p$ for the label $l$, and $r\in(0,1)$ is a random number sampled from the uniform distribution, which decides the location of the synthetic point between $\mathbf{u}_p$ and $\mathbf{v}_p$.

The number of synthetic minority points generated from a cluster is directly proportional to the share of original minority points in that cluster. Therefore, more synthetic minority points will be introduced in the clusters with more original minority points. This is because, we are more confident about adding minority points in a region which originally had comparatively more original minority points. The number of synthetic minority points to be added is computed as follows

\begin{equation}
syn_{lp} = \lceil n_{lp}\times\frac{|maj_{l}|-|min_{l}|}{|min_{l}|} \rceil
\end{equation}

\noindent where $min_{l}$ and $maj_{l}$ are the sets of minority and majority datapoints for the label $l$ respectively. Here $n_{lp}$ is the number of original minority datapoints for label $l$ in cluster $C_p$. This way, the clusters which have more original minority points will be populated with more synthetic minority point.




Following the above steps, once we obtain the synthetic minority set $\mathcal{S}_l$ for the label $l$, the original training dataset $\mathcal{D}$ is appended with $\mathcal{S}_l$ to get an augmented dataset $\mathcal{A}_l$ for each label $l$. This augmented training set, $\mathcal{A}_l$, is used to train a binary classifier model for the corresponding label $l$. The above process is summarised in Algorithm \ref{alg:uclso}.

\section{Experiments}\label{sec:experiments}

We performed a set of experiments to evaluate the effectiveness of the proposed UCLSO method. This section describes the datasets, algorithms, experimental setup, and evaluation processes used for the experiments.
{
\begin{table}
\setlength{\tabcolsep}{5pt} 
\renewcommand{\arraystretch}{1} 
\centering
\caption{Description of datasets\label{tab:datasets}}
\resizebox{\textwidth}{!}{ %

\begin{tabular}{lrrrrrrrrrrr}

\hline 
Dataset & Instances & Inputs & Labels & Type & Cardinality & Density & Distinct  & Proportion of  & \multicolumn{3}{c}{Imbalance Ratio}\tabularnewline
\cline{10-12} 
 &  &  &  &  &  &  & Labelsets & Distinct  & min & max & avg\tabularnewline
 &  &  &  &  &  &  &  &  Labelsets &  &  & \tabularnewline
\hline 
yeast & 2417 & 103 & 13 & numeric & 4.233 & 0.325 & 189 & 0.078 & 1.328 & 12.500 & 2.778\tabularnewline
emotions & 593 & 72 & 6 & numeric & 1.869 & 0.311 & 27 & 0.046 & 1.247 & 3.003 & 2.146\tabularnewline
medical & 978 & 144 & 14 & numeric & 1.075 & 0.077 & 42 & 0.043 & 2.674 & 43.478 & 11.236\tabularnewline
cal500 & 502 & 68 & 124 & numeric & 25.058 & 0.202 & 502 & 1.000 & 1.040 & 24.390 & 3.846\tabularnewline
rcv1-s1 & 6000 & 472 & 42 & numeric & 2.458 & 0.059 & 574 & 0.096 & 3.342 & 49.000 & 24.966\tabularnewline
rcv1-s2 & 6000 & 472 & 39 & numeric & 2.170 & 0.056 & 489 & 0.082 & 3.216 & 47.780 & 26.370\tabularnewline
rcv1-s3 & 6000 & 472 & 39 & numeric & 2.150 & 0.055 & 488 & 0.081 & 3.205 & 49.000 & 26.647\tabularnewline
enron & 1702 & 50 & 24 & nominal & 3.113 & 0.130 & 547 & 0.321 & 1.000 & 43.478 & 5.348\tabularnewline
bibtex & 7395 & 183 & 26 & nominal & 0.934 & 0.036 & 377 & 0.051 & 6.097 & 47.974 & 32.245\tabularnewline
llog & 1460 & 100 & 18 & nominal & 0.851 & 0.047 & 109 & 0.075 & 7.538 & 46.097 & 24.981\tabularnewline
corel5k & 5000 & 499 & 44 & nominal & 2.241 & 0.050 & 1037 & 0.207 & 3.460 & 50.000 & 17.857\tabularnewline
slashdot & 3782 & 53 & 14 & nominal & 1.134 & 0.081 & 118 & 0.031 & 5.464 & 35.714 & 10.989\tabularnewline
\hline 
\end{tabular}
}
\end{table}
}
Several well-known multi-label datasets were selected which are listed in Table \ref{tab:datasets} \footnote{http://mulan.sourceforge.net/datasets-mlc.html}. Here, \emph{instances}, \emph{inputs} and \emph{labels} indicate the total number of datapoints, the number of predictor variables, and the number of potential labels respectively in each dataset. \emph{Type} indicates if the input space is numeric or nominal. \emph{Distinct labelsets} indicates the number of unique combinations of labels. \emph{Cardinality} is the average number of labels per datapoint, and \emph{Density} is achieved by dividing Cardinality by the Labels.

The datasets are modified as recommended in \cite{cocoa,He}. Labels having a very high degree of imbalance (50 or greater) or having too few positive samples (20 in this case) are removed. For text datasets (\emph{medical}, \emph{enron}, \emph{rcv1}, \emph{bibtex}), only the input space features with high degree of document frequencies are retained.

To compare the performance of different approaches, we have selected the label-based macro-averaged F-Score and label-based macro-averaged AUC scores recommended in \cite{cocoa}. For the experiments evaluating the proposed algorithm we have performed a $10 \times 2$ fold cross-validation experiment. The experiment setup and environment was kept identical to Zhang et. al.\cite{cocoa}. For clustering, the number of clusters was set to $5$ for the k-means step of UCLSO. In the classification phase, a set of linear SVM classifiers are used, one for each label.

We compare the performance of UCLSO against several state-of-the-art multi-label classification algorithms -- COCOA \cite{cocoa}, THRSEL \cite{thrsel}, IRUS \cite{irus}, SMOTE-EN \cite{smote}, RML \cite{rml}, and binary relevance (BR), calibrated label ranking (CLR) \cite{Calibrated}, ensemble classifier chains (ECC) \cite{cc} and RAkEL \cite{rakel}. We base our experiments on the experiment presented in Zhang et. al. \cite{cocoa}, and extend the results of that paper by adding the performance of UCLSO.

\section{Results}\label{sec:results}

Tables \ref{tab:fscore} and \ref{tab:auc} shows the label-based macro-average F-Score and label-based macro-averaged AUC results respectively\footnote{Note that results for Table \ref{tab:auc} does not have the results RML \cite{rml} as the implementation does not provide prediction scores.}, along with the relative ranks in brackets (lower ranks are better) of the algorithms compared for each dataset. The last row of both tables indicate the average rank for the algorithms. The best values are highlighted in boldface.

Also, to further analyse the differences between the algorithms, we performed a non-parametric statistical test for a multiple classifier comparison test. Following \cite{garcia2010advanced}, we have performed a Friedman test with Finner $p$-value adjustments, and the critical difference plots from the test results are shown in Figure \ref{fig:cdplot} \footnote{The full result tables in supplementary material: \url{ https://github.com/phoxis/uclso/blob/main/UCLSO_ICONIP2021_Supplementary_Material.pdf}}.


\begin{table}[t]
\centering
\caption{Each cell indicates the averaged \textit{label-based macro-averaged F-Scores} scores (best score in bold) along with the relative rank of the corresponding algorithm in brackets. The last row indicates the overall average ranks.\label{tab:fscore}}
\resizebox{\textwidth}{!}{ %
\setlength{\tabcolsep}{4pt} 
\renewcommand{\arraystretch}{1} 
\begin{tabular}{l r r r r r r r r r r}
    \hline
            & UCLSO     & COCOA     & THRSEL    & IRUS & SMOTE-EN & RML & BR & CLR & ECC & RAkEL \\ 
  \hline
  yeast     & \textbf{0.505} (1) & 0.461 (3) & 0.427 (5\ \ ) & 0.426 (6\ ) & 0.436 (4\ \ ) & 0.471 (2\ \ ) & 0.409 (9\ ) & 0.413 (8\ ) & 0.389 (10\ )  & 0.420 (7) \\ 
  emotions  & 0.658 (2) & \textbf{0.666} (1) & 0.560 (9\ \ ) & 0.622 (5\ ) & 0.575 (8\ \ ) & 0.645 (3\ \ ) & 0.550 (10)  & 0.595 (7\ ) & 0.638 (4\ \ ) & 0.613 (6) \\ 
  medical   & \textbf{0.783} (1) & 0.759 (2) & 0.733 (3.5)   & 0.537 (10)  & 0.700 (8\ \ ) & 0.707 (7\ \ ) & 0.718 (6\ ) & 0.724 (5\ ) & 0.733 (3.5)   & 0.672 (9) \\ 
  cal500    & 0.273 (2) & 0.210 (5) & 0.252 (3\ \ ) & \textbf{0.277} (1\ ) & 0.235 (4\ \ ) & 0.209 (6\ \ ) & 0.169 (8\ ) & 0.081 (10)  & 0.092 (9\ \ ) & 0.193 (7) \\ 
  rcv1-s1   & \textbf{0.443} (1) & 0.364 (3) & 0.292 (5\ \ ) & 0.252 (8\ ) & 0.313 (4\ \ ) & 0.387 (2\ \ ) & 0.285 (6\ ) & 0.227 (9\ ) & 0.192 (10\ )  & 0.272 (7) \\ 
  rcv1-s2   & \textbf{0.432} (1) & 0.342 (3) & 0.275 (5\ \ ) & 0.234 (8\ ) & 0.305 (4\ \ ) & 0.363 (2\ \ ) & 0.272 (6\ ) & 0.226 (9\ ) & 0.173 (10\ )  & 0.263 (7) \\ 
  rcv1-s3   & \textbf{0.480} (1) & 0.339 (3) & 0.275 (5\ \ ) & 0.225 (8\ ) & 0.302 (4\ \ ) & 0.371 (2\ \ ) & 0.271 (6\ ) & 0.211 (9\ ) & 0.163 (10\ )  & 0.257 (7) \\ 
  enron     & \textbf{0.352} (1) & 0.342 (2) & 0.291 (5\ \ ) & 0.293 (4\ ) & 0.266 (8\ \ ) & 0.307 (3\ \ ) & 0.246 (9\ ) & 0.244 (10)  & 0.268 (6\ \ ) & 0.267 (7) \\ 
  bibtex    & \textbf{0.442} (1) & 0.318 (3) & 0.303 (4\ \ ) & 0.253 (8\ ) & 0.283 (5\ \ ) & 0.326 (2\ \ ) & 0.263 (7\ ) & 0.265 (6\ ) & 0.212 (10\ )  & 0.252 (9) \\ 
  llog      & \textbf{0.181} (1) & 0.082 (6) & 0.096 (3\ \ ) & 0.124 (2\ ) & 0.095 (4.5)   & 0.095 (4.5)   & 0.031 (7\ ) & 0.024 (8\ ) & 0.022 (10\ )  & 0.023 (9) \\ 
  corel5k   & 0.209 (2) & 0.196 (3) & 0.146 (4\ \ ) & 0.105 (6\ ) & 0.125 (5\ \ ) & \textbf{0.215} (1\ \ ) & 0.089 (7\ ) & 0.049 (10)  & 0.054 (9\ \ ) & 0.084 (8) \\ 
  slashdot  & \textbf{0.443} (1) & 0.374 (2) & 0.355 (4\ \ ) & 0.257 (10)  & 0.366 (3\ \ ) & 0.343 (5\ \ ) & 0.291 (8\ ) & 0.290 (9\ ) & 0.304 (6\ \ ) & 0.296 (7) \\ 
  \hline
   Avg. rank & \textbf{1.25} & 3.00 & 4.62 & 6.33 & 5.12 & 3.29 & 7.42 & 8.33 & 8.12 & 7.5 \\ 
   \hline

\end{tabular}

}
\end{table}

\begin{table}[t]
\centering
\caption{Each cell indicates the averaged \textit{Label-based macro-averaged AUC} scores (best score in bold) along with the relative rank of the corresponding algorithm in brackets. The last row indicates average ranks.\label{tab:auc}}
\resizebox{\textwidth}{!}{ %
\setlength{\tabcolsep}{4pt} 
\renewcommand{\arraystretch}{1} 

\begin{tabular}{lrrrrrrrrr}
  \hline
             & UCLSO     & COCOA & THRSEL & IRUS & SMOTE-EN & BR & CLR & ECC & RAkEL \\ 
  \hline
  yeast      & 0.666 (3) & \textbf{0.711} (1\ \ ) & 0.576 (8.5) & 0.658 (4\ \ ) & 0.582 (7) & 0.576 (8.5) & 0.650 (5\ \ ) & 0.705 (2\ \ ) & 0.641 (6) \\ 
  emotions   & 0.819 (3) & 0.844 (2\ \ ) & 0.687 (8.5) & 0.802 (4\ \ ) & 0.698 (7) & 0.687 (8.5) & 0.796 (6\ \ ) & \textbf{0.850} (1\ \ ) & 0.797 (5) \\ 
  medical    & \textbf{0.967} (1) & 0.964 (2\ \ ) & 0.869 (7.5) & 0.955 (3.5)   & 0.873 (6) & 0.869 (7.5) & 0.955 (3.5)   & 0.952 (5\ \ ) & 0.856 (9) \\ 
  cal500     & 0.550 (4) & 0.558 (2\ \ ) & 0.509 (8.5) & 0.545 (5\ \ ) & 0.512 (7) & 0.509 (8.5) & \textbf{0.561} (1\ \ ) & 0.557 (3\ \ ) & 0.528 (6) \\ 
  rcv1-s1    & \textbf{0.919} (1) & 0.889 (3\ \ ) & 0.643 (7.5) & 0.882 (4\ \ ) & 0.626 (9) & 0.643 (7.5) & 0.891 (2\ \ ) & 0.881 (5\ \ ) & 0.728 (6) \\ 
  rcv1-s2    & \textbf{0.912} (1) & 0.882 (2.5)   & 0.640 (7.5) & 0.880 (4\ \ ) & 0.622 (9) & 0.640 (7.5) & 0.882 (2.5)   & 0.874 (5\ \ ) & 0.721 (6) \\ 
  rcv1-s3    & \textbf{0.956} (1) & 0.880 (2\ \ ) & 0.633 (7.5) & 0.872 (4.5)   & 0.628 (9) & 0.633 (7.5) & 0.877 (3\ \ ) & 0.872 (4.5)   & 0.718 (6) \\ 
  enron      & 0.719 (5) & \textbf{0.752} (1\ \ ) & 0.597 (8.5) & 0.738 (3\ \ ) & 0.619 (7) & 0.597 (8.5) & 0.720 (4\ \ ) & 0.750 (2\ \ ) & 0.650 (6) \\ 
  bibtex     & 0.844 (4) & 0.877 (2\ \ ) & 0.673 (8.5) & \textbf{0.894} (1\ \ ) & 0.706 (6) & 0.673 (8.5) & 0.811 (5\ \ ) & 0.873 (3\ \ ) & 0.696 (7) \\ 
  llog       & \textbf{0.721} (1) & 0.663 (4\ \ ) & 0.518 (7.5) & 0.676 (2\ \ ) & 0.561 (6) & 0.518 (7.5) & 0.612 (5\ \ ) & 0.673 (3\ \ ) & 0.514 (9) \\ 
  corel5k    & 0.695 (4) & 0.718 (3\ \ ) & 0.559 (7.5) & 0.687 (5\ \ ) & 0.596 (6) & 0.559 (7.5) & \textbf{0.740} (1\ \ ) & 0.723 (2\ \ ) & 0.552 (9) \\ 
  slashdot   & \textbf{0.806} (1) & 0.774 (2\ \ ) & 0.632 (8.5) & 0.753 (4\ \ ) & 0.714 (6) & 0.632 (8.5) & 0.742 (5\ \ ) & 0.765 (3\ \ ) & 0.638 (7) \\ 
  \hline
  Avg. ranks & 2.42 & \textbf{2.21} & 8.00 & 3.67 & 7.08 & 8.00 & 3.58 & 3.21 & 6.83 \\ 
   \hline
    
\end{tabular}
}
\end{table}

Table \ref{tab:fscore} clearly shows that the overall performance of the proposed UCLSO algorithm is better than all the other algorithms, attaining the best average rank of 1.25. The second best rank is attained by COCOA (avg. rank 3). Also, the proposed method UCLSO achieved much better performance than the other approaches for many datasets and attained the top rank for nine of the datasets, and on the remaining three datasets it attained the second rank. these results also show that methods that attempt to explicitly consider the label imbalance issue perform better than those that do not. The other algorithms which specifically address label imbalance attained the following order: RML (avg. rank 3.29), THRSEL (avg. rank 4.62), SMOTE-EN (avg. rank 5.12) and IRUS (arg. rank 6.33). The algorithms which do not consider the label imbalances like BR (avg. rank 7.42), RAkEL (avg. rank 7.5), ECC (avg. rank 8.12), and CLR (avg. rank 8.33) all performed poorly. 

Multiple classifier comparison results in Figure \ref{fig:friedman_f1} show that when UCLSO is compared with other algorithms, except for COCOA and RML,  the null hypothesis can be rejected with a significance level of $\alpha = 0.05$. Therefore, based on the statistical test, UCLSO is significantly better than  the other algorithms, except COCOA and RML.

\begin{figure}
\centering\makebox[\textwidth]{
\subfloat[\emph{label-based macro-averaged F-Score}\label{fig:friedman_f1}]{\includegraphics[width=0.5\textwidth]{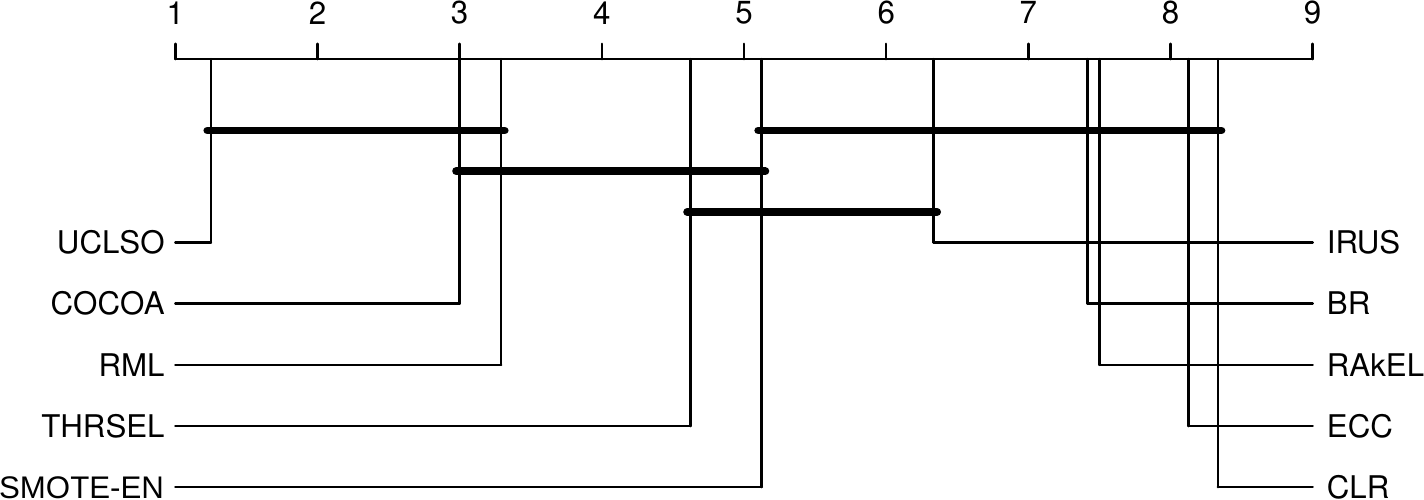}}
\subfloat[\emph{label-based macro-averaged AUC}\label{fig:friedman_auc}]{\includegraphics[width=0.5\textwidth]{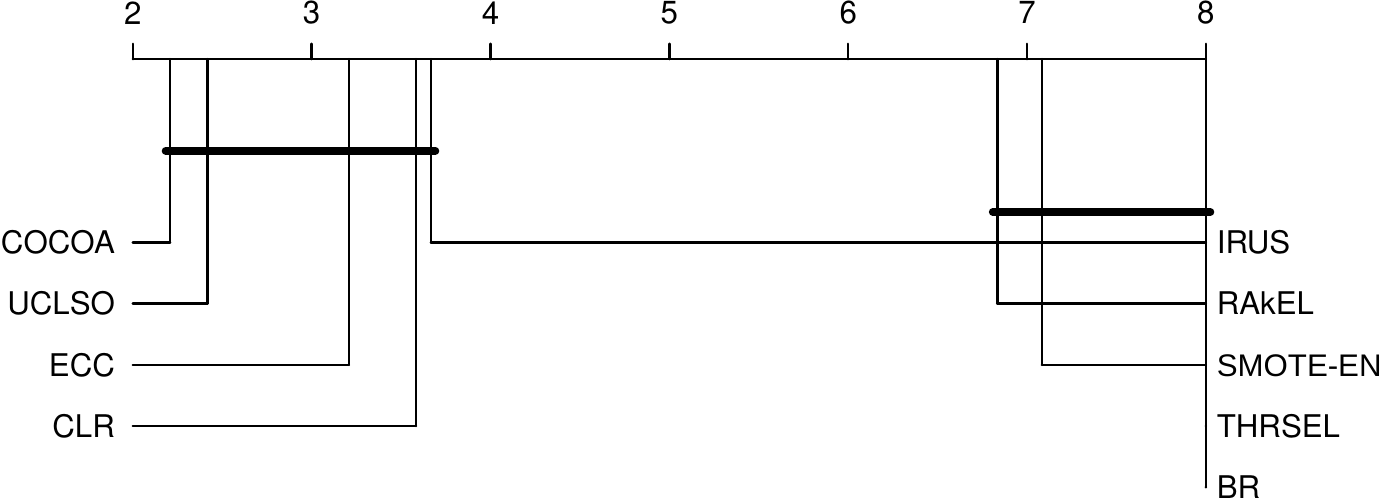}}
}\caption{Critical difference plots. The scale indicates the average ranks. The methods which are not connected with the horizontal lines are significantly different with a significance level of $\alpha = 0.05$.}\label{fig:cdplot}
\end{figure}

Table \ref{tab:auc} shows the label-based macro-averaged AUC scores, which shows that proposed method UCLSO was able to attain the second best average rank of 2.42, being very close to COCOA attaining the best rank of 2.21. Interestingly UCLSO attained more rank ones (six) than COCOA (two rank ones). Also, interestingly ECC was able to perform better than UCLSO in six of the datasets, but was able to perform better in nine datasets when compared to COCOA. It is also interesting to notice that ECC and CLR had higherrankings for the label-based macro-averaged AUC metric than for macro-averaged F-Scores. It seems that a simple BR still performed poorly. As ECC and CLR takes label associations into consideration in a binary relevance and ranking fashion, respectively, it helped improve the comparative performances. RAkEL, on the other hand, taking label associations into account is sensitive on the label subset size (value of k) and the specific combination, which can lead to an even higher degree of imbalance. The difference in the results of the label-based macro-average AUC compared to the F-Score also indicates the importance of thresholding the predictions when deciding the relevance of a certain label.

Multiple classifier comparison results show in Figure \ref{fig:friedman_auc} that when UCLSO is compared with others, the null hypothesis could not be rejected for COCOA, ECC, CLR and IRUS in this case with a significance level of $\alpha = 0.05$. Although, UCLSO performed significantly better than RAkEL, SMOTE-ML, THRSEL and BR. Overall, the experiments demonstrate the effectiveness of the proposed method UCLSO, as it outperforms the compared state of the art algorithms in almost all  cases.

\section{Conclusion and Future Work}\label{sec:conclusion}

In this work we have proposed an algorithm to address the class imbalance of labels in multi-label classification problems. The proposed algorithm, Unsupervised Clustering and Label-Specific data Oversampling (UCLSO), oversamples label-specific minority datapoints in a multi-label problem to balance the sizes of the majority and the minority classes of each label. The oversampling of the minority classes for each label is done in a way such that more minority class samples are generated in regions (or clusters) where the density of minority points is high. This avoids the introduction of minority datapoints in majority regions in the input space. The number of samples introduced per cluster also depends on the share of the minority class for that cluster.

An experiment with 12 well-known multi-label datasets and other state of the art algorithms demonstrates the efficacy of UCLSO with respect to label-based macro-averaged F-Score. UCLSO attained the best average rank and the degree of its improvement over existing approaches was significant. This shows that UCLSO has successfully improved the classification of imbalanced multi-label data. In future, we would specifically like to incorporate some imbalance informed clustering to extend our scheme. Moreover, it would be interesting to amalgamate the oversampling technique with label associated learning, another key component of multi-label data.

\bibliographystyle{splncs04}
\bibliography{cbosampml}

\end{document}